\documentclass[10pt,twocolumn,letterpaper]{article}
\usepackage[pagenumbers]{cvpr}
\usepackage{graphicx}
\AtBeginDocument{%
  \crefname{figure}{Figure}{Figures}%
  \Crefname{figure}{Figure}{Figures}}
\usepackage{amsmath,amssymb}
\usepackage{booktabs}
\usepackage{tabularx}
\usepackage{multirow}
\usepackage[table]{xcolor}
\usepackage{wrapfig,pifont}
\newcommand{\papertitle}{Recognizing Events Is Not Enough: Why Streaming VLMs Struggle in Compositional Instruction Triggering}

\newcommand{\na}{--}

\usepackage{float}
\usepackage{balance}
\usepackage{fvextra}
\usepackage{CJKutf8}
\DeclareUnicodeCharacter{2192}{\ensuremath{\rightarrow}}
\definecolor{cvprblue}{rgb}{0.21,0.49,0.74}
\usepackage[pagebackref,breaklinks,colorlinks,allcolors=cvprblue]{hyperref}
\def\confName{CVPR}
\def\confYear{2027}

\title{\papertitle}
\author{%
\parbox{\dimexpr\textwidth-2\tabcolsep\relax}{\centering
Yehua Huang$^{1}$, Penglei Sun$^{1}$, Xiang Li$^{1}$, Jinhao Zhou$^{2}$, Jiawen Wen$^{1}$\\
Qingqing Xiao$^{3}$, Zihan Yang$^{3}$, Wenqi Wang$^{3}$, Haolan Chen$^{3}$, Yu Xu$^{3}$, Xiaowen Chu$^{1,4}$\\[4pt]
\small
$^{1}$The Hong Kong University of Science and Technology (Guangzhou) (hkust-gz.edu.cn)\\
$^{2}$Tsinghua University (tsinghua.edu.cn)\\
$^{3}$Tencent (tencent.com)\\
$^{4}$The Hong Kong University of Science and Technology (ust.hk)
}}
\hypersetup{pdftitle={Recognizing Events Is Not Enough: Why Streaming VLMs Struggle in Compositional Instruction Triggering},pdfauthor={Yehua Huang, Penglei Sun, Xiang Li, Jinhao Zhou, Jiawen Wen, Qingqing Xiao, Zihan Yang, Wenqi Wang, Haolan Chen, Yu Xu, Xiaowen Chu}}
\begin{document}
\maketitle
\begin{abstract}
Recognizing individual events does not guarantee correct execution of compositional instructions in streaming video. 
For example, given the request \textit{``Notify me when the pot boils after someone turns on the stove and leaves the kitchen''}, existing models may recognize the individual events yet notify too early or miss the intended moment, failing to track how their temporal conditions are satisfied over time.
To evaluate this gap, we introduce CoTriggerBench, which composes annotated video events into temporally constrained instructions and verifies notification timing and silence through executable event plans and source-event replay.
To address this challenge, we propose a real-time VLM agent, TEMPI.
TEMPI decomposes instructions into atomic event queries and temporal constraints, and uses Scenario Memory to continuously maintain task progress as event evidence arrives, allowing earlier observations to support subsequent decisions beyond the detector's visual window.
Even when no notification is required, relevant events continue to update the scenario state.
On CoTriggerBench, TEMPI achieves 44.4 \% instruction-level success, outperforming the strongest closed-source and fine-tuned open-source VLM baselines by 15.6 (Qwen3.8-Omni-Flash-Realtime) and 18.1 (MOSS-VL-Realtime) percentage points, respectively.
To support efficient on-device streaming inference, we further optimize the streaming inference pipeline by overlapping partial visual encoding and LLM prefix prefill with frame arrival. On an 8 GB RTX 5060, our optimized detector implementation achieves a 4.90$\times$ speedup.
\end{abstract}

\begin{figure}[!t]
\centering
\begin{subfigure}[t]{0.58\linewidth}
\vspace{0pt}
\includegraphics[width=\linewidth]{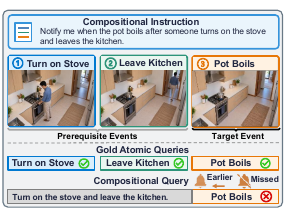}
\caption{Failure case study}\label{fig:scenario-memory-case-failure}
\end{subfigure}%
\hspace{0.012\linewidth}%
\begin{subfigure}[t]{0.408\linewidth}
\vspace{0pt}
\includegraphics[width=\linewidth]{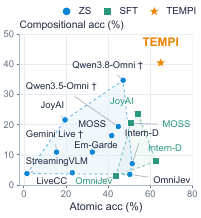}
\caption{Model comparison}\label{fig:scenario-memory-case-scatter}
\end{subfigure}
\caption{\textbf{Atomic and compositional triggering.}
Figure~\textbf{\subref{fig:scenario-memory-case-failure}} shows that existing models succeed with atomic queries and explicit execution but fail on the compositional instruction.
Figure~\textbf{\subref{fig:scenario-memory-case-scatter}} reports request-level accuracy rates in atomic and composite triggering events.
Most models achieve much higher trigger accuracy in atomic events than compositional events.
$\dagger$ denotes closed models.}
\label{fig:scenario-memory-case}
\end{figure}

\section{Introduction}
\label{sec:intro}

Streaming video assistants continuously observe incoming video and respond to users based on their queries.
They can support everyday tasks by monitoring relevant situations and providing timely assistance, reducing the need for users to watch continuously.
In practice, users express their needs through natural-language instructions that often include prerequisites.
For example, a user may ask, \textit{``Notify me when the pot boils after someone turns on the stove and leaves the kitchen''} (Figure~\ref{fig:scenario-memory-case-failure}).
Detecting the pot boiling alone is insufficient.
The assistant must also verify that the stove was lit and a person left the kitchen before the pot boils.
Premature or missed notifications undermine the purpose of delegation.
Motivated by this need, we study \textbf{compositional event triggering}: determining when user-specified combinations of events and temporal constraints are satisfied over a video stream, and issuing notifications at the appropriate moment using only observations available so far.

A key challenge in compositional event triggering is maintaining instruction progress across video observations, even when no notification is required. 
In the stove example, we refer to turning on the stove and leaving the kitchen as prerequisite events, and the pot boiling as the target event.
As prerequisite events are observed, the assistant must update its record of satisfied conditions without issuing a notification. When the target event occurs, it must determine whether the accumulated evidence satisfies the instruction’s prerequisites and temporal constraints.
Existing approaches learn response readiness~\cite{mmduet,streamready} or retain historical evidence and structured task information~\cite{onestreamer,vinci2,emgarde,proact}, but these capabilities do not necessarily ensure correct execution of compositional instructions. 
As shown in Figure~\ref{fig:scenario-memory-case-scatter}, fine-tuning on task-specific event data~\cite{perceptiontest} improves atomic-event accuracy, while gains in compositional triggering remain limited. Several fine-tuned VLMs~\cite{joyai,moss,intern} achieve over $50 \%$ atomic accuracy but remain below $25 \%$ compositional accuracy.
This gap indicates that recognizing individual events is insufficient for executing compositional instructions.

\begin{figure*}[!t]
\centering
\includegraphics[width=\textwidth]{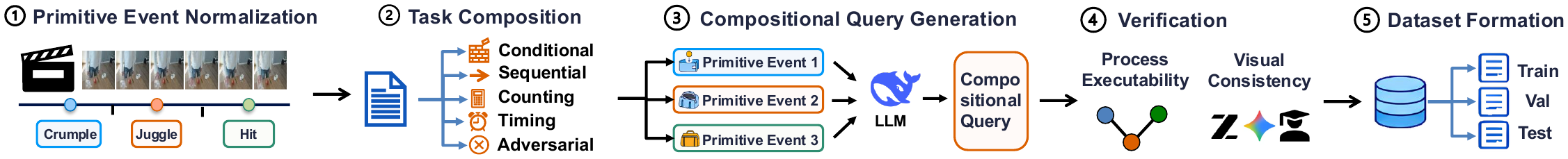}
\caption{\textbf{CoTriggerBench construction pipeline.} The full construction pipeline comprises event normalization, task composition, query generation, verification, and dataset formation.}
\label{fig:teaser}
\end{figure*}

To evaluate the gap between atomic event recognition and compositional instruction execution, we introduce \textbf{CoTriggerBench}. 
Using event annotations from Perception Test~\cite{perceptiontest}, we define five task types: \emph{conditional}, \emph{sequential}, \emph{counting}, \emph{timing}, and \emph{adversarial}, covering prerequisites, event ordering, occurrence counts, delays, and silence when the specified conditions are not satisfied. 
We construct the tasks through event composition, natural-language instruction generation, and consistency verification, with expected outcomes validated through deterministic event replay (Figure~\ref{fig:teaser}).
The test set contains 448 instructions across 96 videos, including 82 adversarial cases, evaluating notification timing and correct silence.

We introduce \textbf{TEMPI} (\emph{Temporal Execution and Monitoring of Planned Instructions}), a streaming video agent for compositional event triggering that separates event detection from instruction execution. 
TEMPI maintains instruction progress in \emph{Scenario Memory}, consisting of a \emph{Scenario Plan} and a \emph{Scenario State}.
Specifically, a 4B planner decomposes each instruction into atomic event queries and their compositional constraints, while a 2B detector extracts timestamped event evidence from the video stream. 
During execution, TEMPI updates the scenario state based on the event evidence and compositional constraints, tracking satisfied conditions, occurrence counts, and timing information to determine when to issue notifications. 
Even when no notification is required, relevant events continue to update the scenario state, allowing earlier observations to inform subsequent decisions after they leave the visual window.

On CoTriggerBench, TEMPI achieves the SOTA instruction-level success rate in compositional instruction execution (44.4\%), outperforming existing real-time VLMs such as Qwen3.8-Omni-Flash-Realtime (28.8\%) and Gemini 3.8 Live (20.5\%), as well as MOSS-VL-Realtime (11B, 26.3\%) and JoyAI-VL-Interaction (8B, 26.1\%), which are fine-tuned on the same dataset as TEMPI.
Further experiments demonstrate that these improvements stem not only from event perception but also from explicit instruction decomposition and stateful execution. 
The TEMPI framework also improves compositional instruction execution across different perception backends. For example, integrating TEMPI improves instruction-level success for fine-tuned JoyAI from 26.1 \% to 39.1\% and Intern-Decision from 18.3\% to 43.1\%, demonstrating its applicability across diverse perception models.
On a single H800, TEMPI achieves a median GPU execution time of only 61.0 ms per query/window, compared with 240.3 ms for JoyAI, representing 3.94$\times$ faster GPU execution.

To further support efficient on-device streaming inference, we propose a streaming early-execution strategy.
Instead of waiting for all four frames in a decision window, we initiate partial ViT encoding and LLM prefix prefill after the first two frames arrive.
Once the remaining frames become available, inference continues with suffix prefill by reusing the cached prefix KV states, thereby reducing decision latency after the final frame arrives. 
On an 8 GB RTX 5060, the optimized NVFP4 implementation achieves up to a 4.90$\times$ speedup in post-window decision latency, while nearly preserving instruction-level success (43.8\% vs. 44.4\%), demonstrating TEMPI's potential for efficient streaming inference under edge deployment.

Our contributions are as follows:
\begin{itemize}
\item We formulate compositional instruction triggering in streaming video and introduce CoTriggerBench.
CoTriggerBench evaluates instructions involving prerequisites, event ordering, occurrence counts, and delays. 
By assessing both timely notifications and correct silence, CoTriggerBench reveals the gap between event recognition and compositional instruction execution in streaming VLMs.

\item We propose TEMPI, a compositional-triggering agentic framework that integrates instruction planning, event detection, and stateful execution to track task progress and determine when to issue notifications.
Specifically, TEMPI achieves the highest success rate on CoTriggerBench (44.4\%), outperforming the strongest baseline (28.8\%).

\item We develop an efficient streaming-aware inference pipeline for edge deployment. Instead of waiting for the complete video window, our pipeline overlaps early ViT encoding and LLM prefix prefill with streaming frame arrival, reusing cached prefix KV states to minimize post-window decision latency. On an 8 GB RTX 5060, our optimized implementation achieves a 4.90$\times$ speedup over native inference for three queries, while preserving effectiveness.

\end{itemize}

\section{Related Work}
\label{sec:related}
\paragraph{Benchmarks for proactive interaction.}
Existing benchmarks evaluate diverse aspects of proactive video interaction. StreamingBench~\cite{streamingbench} and OVO-Bench~\cite{ovobench} cover online video understanding; OmniMMI~\cite{omnimmi}, ProactiveVideoQA~\cite{proactivevideoqa}, LiveProBench~\cite{liveprobench}, and OmniPro~\cite{omnipro} assess alerting, response timing, silence, counting, and monitoring, while IPIBench~\cite{ipibench} focuses on dynamic requests. In contrast, CoTriggerBench evaluates single-turn instructions with explicit temporal constraints, including prerequisites, event ordering, occurrence counts, and event-relative delays, requiring timely notifications when conditions are satisfied and silence otherwise.

\paragraph{Proactive streaming models.}
Executing temporally constrained instructions requires maintaining task progress beyond local visual observations. Existing streaming models retain visual context through compact or hierarchical memory~\cite{streamingvlm,ovbench,onestreamer}, regulate responses through readiness estimation or inference gating~\cite{mmduet,streambridge,streammind,streamready}, and support continuous interaction or lightweight decisions~\cite{joyai,moss,intern,omnijev}. Other approaches match event cues~\cite{emgarde}, anticipate relevant observations~\cite{streamagent}, or leverage task graphs for proactive actions~\cite{proact}, while structured programs compose perception and execution~\cite{visprog,codeaspolicies}. In contrast, TEMPI explicitly represents prerequisites, ordering, counts, and delays through Scenario Memory, maintaining persistent execution state from interchangeable event detectors. 

\section{Benchmark}
\label{sec:benchmark}
We introduce CoTriggerBench, a benchmark for compositional event triggering across five notification task types grounded in annotated video events.

\subsection{Task Definition}

Given a user instruction $q$ and a streaming video $V$, a streaming VLM $f_\theta(\cdot)$ produces an output at time $t$ based on the current and past video observations $V_{\leq t}$:
\begin{equation}
o_t=f_\theta(q,V_{\leq t}).
\label{eq:stream-output}
\end{equation}
Here, $q$ is a natural language instruction that specifies logical and temporal constraints, and $o_t$ represents either a generated notification or silence.

\begin{figure}[t]
\centering
\includegraphics[width=0.85\linewidth]{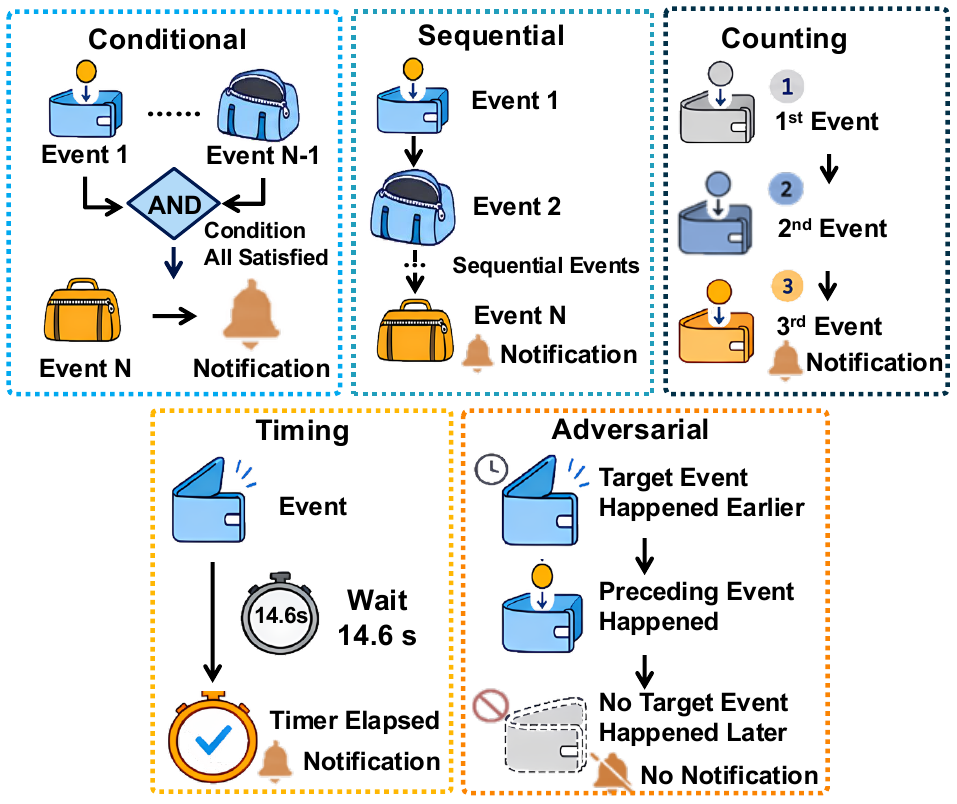}
\caption{\textbf{CoTriggerBench task types.} This figure shows event timelines and notification outcomes for the five task types.}
\label{fig:task-taxonomy}
\end{figure}

\paragraph{Task Types.}
Let $A_i^{t_i}$ denote that event $A_i$ occurs at time $t_i$, where $t$ is the current decision time.
We define five task types to evaluate whether streaming VLMs can follow complex instructions involving event conditions, temporal order, counting, and timing (Figure~\ref{fig:task-taxonomy}).

\noindent$\bullet$ \textbf{Conditional.}
A notification is triggered when the target event $A_n$ occurs after all required conditions have been satisfied, regardless of their relative order, defined as $q= (\bigwedge_{i=1}^{n-1} A_i^{t_i}) \prec A_n^t$.

\noindent$\bullet$ \textbf{Sequential.}
A notification is triggered when events occur in the specified order, with the current event completing the sequence, defined as $q= A_1^{t_1}\prec A_2^{t_2}\prec\cdots\prec A_n^t$.

\noindent$\bullet$ \textbf{Counting.}
A notification is triggered at the $k$-th distinct occurrence of event $A$, defined as $q= A^t\land[N_A(t)=k]$, where $N_A(t)$ denotes the number of distinct occurrences of $A$ up to time $t$.

\noindent$\bullet$ \textbf{Timing.}
A notification is triggered when a specified time interval $\Delta$ has elapsed since a reference event $A$, defined as $q= A^{t_A}\land[t-t_A=\Delta]$, where $t_A$ denotes the time of the reference event. A small timing tolerance is allowed during evaluation.

\noindent$\bullet$ \textbf{Adversarial.}
Adversarial tasks pair valid instructions with videos that do not satisfy their constraints, requiring the model to remain silent even when some relevant events occur.
For example, a sequential instruction $q= A_1^{t_1}\prec A_2^{t_2}$ may be paired with a video satisfying $\neg(A_1^{t_1}\prec A_2^{t_2})$. Similar negative cases can be constructed for the other tasks.

\subsection{Benchmark Construction}
\paragraph{Event normalization.}
Perception Test~\cite{perceptiontest} provides temporally localized action segments linked to object tracks, with action labels, object identities, and start/end timestamps. Following Figure~\ref{fig:teaser}, we associate each action with its annotated objects, normalize action--object categories, and convert timestamps to seconds while preserving source boundaries and identifiers. We define each resulting record, comprising an action, its associated objects, and a temporal interval, as a \emph{primitive event}.

\begin{wraptable}{r}{0.58\columnwidth}
\centering
\captionsetup{width=\linewidth}
\caption{\textbf{Benchmark split statistics.} Task rows count instructions.}
\label{tab:dataset}
\footnotesize
\setlength{\tabcolsep}{2pt}
\begin{tabular*}{\linewidth}{@{\extracolsep{\fill}}lrrr@{}}
\toprule
Task type & \multicolumn{1}{c}{Train} & \multicolumn{1}{c}{Val.} & \multicolumn{1}{c}{Test}\\
\midrule
Conditional & 9,270 & 95 & 95\\
Sequential & 4,410 & 90 & 96\\
Counting & 3,912 & 87 & 84\\
Timing & 3,806 & 91 & 91\\
Adversarial & 7,457 & 88 & 82\\
\midrule
Total Instructions & 28,855 & 451 & 448\\
Unique Videos & 7,103 & 96 & 96\\
\bottomrule
\end{tabular*}
\end{wraptable}

\paragraph{Task composition.}
Within each video, type-specific rules enumerate task candidates from primitive events. Distinct, nonoverlapping event pairs and ordered triples form conditional and sequential candidates. Repeated occurrences of the same action--object category form counting candidates. Timing candidates attach a duration to an anchor, with timer expiry constrained to the video duration. Adversarial candidates pair an earlier target with a later prerequisite and require no annotated target recurrence afterwards. Temporal-gap and duration checks constrain event selection. Each candidate records event roles, dependencies, any count or delay, and the expected outcome.

\paragraph{Query generation.}
We use DeepSeek-V4.1-Flash~\cite{deepseek41flash} to express structured task candidates as natural-language notification requests. The prompt specifies the task type, primitive-event descriptions, dependency order, and any count or delay. The prompt is provided in the supplementary material. The resulting instruction states the requested trigger condition with explicit temporal requirements and object roles. We also derive canonical, subject-free atomic queries for the constituent primitive events.

\paragraph{Verification.}
GLM-5.3-Flash~\cite{glm53flash} and Gemini 3.8 Flash~\cite{gemini38flash} check the visual grounding of event descriptions and object roles against video evidence. Human annotators additionally inspect all test samples for visual alignment, the naturalness of compositional queries, and consistency between each query and its constituent subgoals. To verify executability, we deterministically compile candidates into gold plans and validate event references, dependency graphs, and count/delay parameters. We check instruction--plan consistency and replay source event intervals using the task-state update rules to verify that the resulting notification or silence matches the expected outcome.

\paragraph{Dataset formation.}

Table~\ref{tab:dataset} summarizes the video-disjoint splits. Training retains the inherited source instruction pool and supplies atomic-event supervision and compiler examples. Validation selects thresholds and policies, and each frozen configuration is evaluated once on test. We exclude evaluation tasks with ambiguous event/object references, unresolved event roles, or invalid plan expansion because they lack an unambiguous, executable ground truth. The retained evaluation sets contain 451 validation tasks and 448 test tasks. Test comprises 366 positive and 82 adversarial tasks over 96 videos.

\begin{figure*}[!t]
\centering
\includegraphics[width=\textwidth]{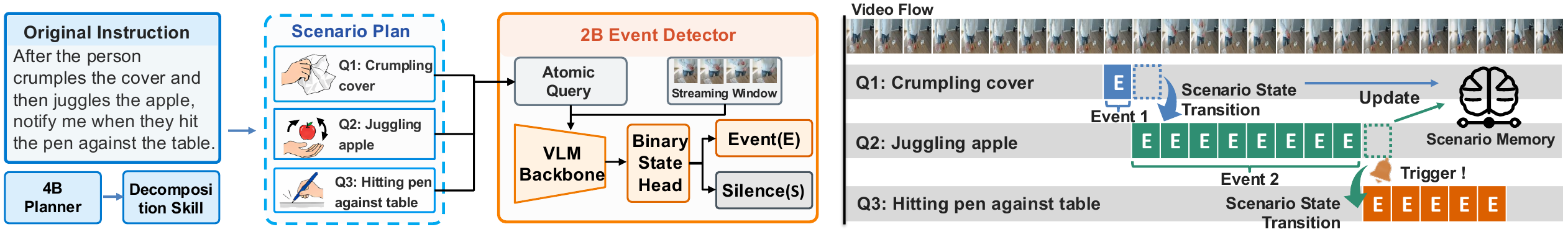}
\caption{\textbf{TEMPI architecture and streaming execution.} A skill-guided TEMPI planner decomposes the instruction into atomic queries and compositional constraints to form the scenario plan. The detector supplies event/silence evidence from recent frames, and the TEMPI agent writes scenario state updates into scenario memory, activates dependent steps, and issues a notification once all constraints are satisfied. External backends can supply the same event interface.}
\label{fig:method}
\end{figure*}

\subsection{Evaluation Protocol}
\paragraph{Trigger timing.}
For a positive task with source target interval $[s,e]$, the first notification is accepted within
\begin{equation}
\mathcal{W}(q)=[s-\delta,e+\delta], \qquad \delta=1.6\ {\rm s}
\label{eq:window}
\end{equation}
For timing tasks, we shift the anchor interval by the requested delay before applying Eq.~\eqref{eq:window}. These intervals are independent of model sampling rates and discretized stream labels.

%

\paragraph{Metrics.}
Let $\mathcal{P}$ and $\mathcal{F}$ denote positive and adversarial tasks. Let $\hat{t}(q)$ be the first notification time, with $\hat{t}(q)=+\infty$ if none is issued. Define $H(q)=\mathbf{1}[\hat{t}(q)\in\mathcal{W}(q)]$, and let $Z(q)$ indicate valid output with no notification. Trigger success (TS), adversarial silence (AS), and strict success are
\par\nobreak
\begingroup
\setlength{\abovedisplayskip}{5pt}
\setlength{\belowdisplayskip}{5pt}
\setlength{\abovedisplayshortskip}{5pt}
\setlength{\belowdisplayshortskip}{5pt}
\noindent
\begin{minipage}[c]{0.48\linewidth}
\begin{equation}
\makebox[\dimexpr\linewidth-2.8em\relax][l]{\hspace*{-1em}$\displaystyle
\makebox[2.8em][r]{$\mathrm{TS}$}=\frac{\sum_{q\in\mathcal{P}}H(q)}{|\mathcal{P}|}$}
\label{eq:ts}
\end{equation}
\end{minipage}\hfill
\begin{minipage}[c]{0.48\linewidth}
\begin{equation}
\makebox[\dimexpr\linewidth-2.8em\relax][l]{\hspace*{-1em}$\displaystyle
\makebox[2.8em][r]{$\mathrm{AS}$}=\frac{\sum_{q\in\mathcal{F}}Z(q)}{|\mathcal{F}|}$}
\label{eq:as}
\end{equation}
\end{minipage}
\begin{equation}
\mathrm{Strict}=\frac{\sum_{q\in\mathcal{P}}H(q)+\sum_{q\in\mathcal{F}}Z(q)}
{|\mathcal{P}|+|\mathcal{F}|}
\label{eq:strict}
\end{equation}
\endgroup

TS is measured on conditional, sequential, counting, and timing tasks that require a notification. It measures whether the first notification falls within the acceptance window for a compositional instruction.
AS is measured only on adversarial tasks, which require silence because their constraints are unmet.
Strict success is the primary metric and reports the overall instruction-level success rate across both groups according to the TS and AS criteria above.
Furthermore, we evaluate single-event detections by measuring all preceding events and triggering events under compositional instructions with the same tolerance and report event-level precision, recall, and F1.


\section{Methodology}
\label{sec:method}
We propose TEMPI, a streaming video agent that separates event perception from compositional instruction execution through Scenario Memory (Figure~\ref{fig:method}). A planner decomposes each instruction into atomic event queries and temporal constraints, forming a Scenario Plan that guides event detection. As event evidence arrives, TEMPI updates the Scenario State and issues a notification once all constraints are satisfied.

\subsection{Task-Conditioned Scenario Memory}
\label{sec:compiler}
\label{sec:scenario-planning}

Unlike streaming models such as JoyAI~\cite{joyai} and MOSS~\cite{moss}, which retain both current and historical visual information within a limited visual-token budget, TEMPI maintains compositional constraints and task progress in Scenario Memory outside the detector's visual context, allowing the detector to focus on local event perception. Specifically, the detector processes non-overlapping windows of four consecutive frames, while Scenario Memory preserves task progress across windows.

Scenario Memory consists of a fixed Scenario Plan and an evolving Scenario State. Guided by a fixed decomposition skill, a planner using Qwen3.5-4B~\cite{qwen35} constructs the plan once per instruction, specifying atomic event queries and compositional constraints, including task types, step dependencies, event boundaries, occurrence counts, and delays. The skill specifies the output schema, task-type selection rules, and examples for extracting atomic queries while preserving action and object semantics. The full planner skill is provided in the supplementary material. The Scenario State records step activation and completion, event counts, and timestamps, and is continuously updated using event evidence extracted from the current video window.

\subsection{Query-Conditioned Event Detection}
\label{sec:detector}
Given a four-frame window $v$ and an atomic event query $q$, the Event Detector predicts whether the queried event is present.
We use Qwen3-VL-2B-Instruct~\cite{qwen3vl} with a two-class linear head applied to the final hidden state. 
The head produces \emph{event} and \emph{silence} logits, where silence denotes absence of the queried event even when other actions occur. 
We train the language backbone and decision head while freezing the vision encoder, visual merger, and shared embeddings.
We define the event margin as $m(v,q)=z_{\rm event}-z_{\rm silence}$ and threshold it at inference.

To distinguish the queried event from other visible actions, we combine background windows and negative windows reliably away from annotated events with query replacements on the same visual window. 
Replacement queries are selected for semantic overlap from the same or other videos and supervised only with valid negative evidence. 
For densely overlapping annotations, we disable same-video replacements and background negatives.

We train the detector with binary state supervision and query contrast. The state loss applies weighted cross-entropy to known event/silence labels, while query contrast compares a positive query with valid replacement negatives on the same window. 
For a positive query $q^+$ and known-negative queries $\mathcal{Q}^-$, define $\mathcal{Q}=\{q^+\}\cup\mathcal{Q}^-$ and $s_q=m(v,q)/\tau$. The objective is
\begin{align}
\mathcal{L}_{\rm qc}
&=-\log\frac{\exp(s_{q^+})}
{\sum_{q\in\mathcal{Q}}\exp(s_q)},\\
\mathcal{L}&=\mathcal{L}_{\rm state}+\lambda\mathcal{L}_{\rm qc},
\label{eq:loss}
\end{align}
where $\tau=1$ and $\lambda=0.25$. 
Query contrast applies only when the positive query is labeled as an event and at least one negative query has valid supervision. 
It penalizes assigning a higher event margin to an irrelevant query than to the supported query, making query discrimination an explicit training objective.

\subsection{Memory-Guided Scenario Execution}
\label{sec:scenario-execution}
As shown in Figure~\ref{fig:method}, upon receiving an instruction, TEMPI invokes its planner once and stores the resulting Scenario Plan of atomic queries and constraints in Scenario Memory.
For each four-frame window, the detector evaluates the atomic queries and returns timestamped event/silence predictions. 
TEMPI groups consecutive positive decisions into an event run and transitions between event and silence identify its onset and offset, allowing TEMPI to determine event completion. 
Completed events update Scenario State, which is maintained in Scenario Memory to track task progress and evaluate constraints against the fixed Scenario Plan. 
Once all prerequisites are satisfied, TEMPI submits the final target query to the detector. 
A valid event prediction for the target triggers a single notification and completes the instruction. Otherwise, TEMPI continues monitoring and updating the state without notifying the user.


\section{Experiments}
\label{sec:experiments}
\begin{table*}[!t]
\centering
\caption{\textbf{Compositional triggering and atomic event detection.} Strict, TS, and AS denote overall instruction success, positive-task trigger success, and adversarial silence. These rates, precision (P), recall (R), and their harmonic mean (F1) are reported as percentages. MAE is the mean absolute distance from the first notification to the unextended target interval, in seconds; FP/min counts false event runs per video minute. Params are in billions (B).
For Em-Garde, 2B+7B denotes its embedding and proposal/response models, respectively. For TEMPI, 2B+4B denotes our detector and planner, respectively.
The best result of each metric is shown in bold.}
\label{tab:paired}
\label{tab:events}
\small
\setlength{\tabcolsep}{2pt}
\begin{tabular*}{\textwidth}{@{\extracolsep{\fill}}lcrrrr@{\hspace{10pt}}rrrr@{}}
\toprule
\multirow{2}{*}{Model} & \multicolumn{1}{c}{\multirow{2}{*}{Params}} & \multicolumn{4}{c}{Compositional Triggering} & \multicolumn{4}{c}{Atomic Event Detection}  \\
\cmidrule(lr){3-6}\cmidrule(lr){7-10}
 &  & \multicolumn{1}{c}{Strict (\%) $\uparrow$} & \multicolumn{1}{c}{TS (\%) $\uparrow$} & \multicolumn{1}{c}{AS (\%) $\uparrow$} & \multicolumn{1}{c}{MAE (s) $\downarrow$} & \multicolumn{1}{c}{F1 (\%) $\uparrow$} & \multicolumn{1}{c}{P (\%) $\uparrow$} & \multicolumn{1}{c}{R (\%) $\uparrow$} & \multicolumn{1}{c}{FP/min $\downarrow$} \\
\midrule
\multicolumn{10}{@{}l}{\textbf{Zero-shot}} \\
JoyAI-VL-Interaction~\cite{joyai} & 8B & 19.2 & 21.6 & 8.5 & 6.84 & 21.3 & 15.8 & 32.8 & 5.62 \\
\addlinespace[2pt]
MOSS-VL-Realtime~\cite{moss} & 11B & 18.3 & 16.4 & 26.8 & 5.10 & 33.3 & 25.1 & 49.7 & 4.76 \\
\addlinespace[2pt]
StreamingVLM~\cite{streamingvlm} & 7B & 15.6 & 3.8 & 68.3 & 6.39 & 3.0 & 27.6 & 1.6 & \textbf{0.14} \\
\addlinespace[2pt]
LiveCC-7B-Instruct~\cite{livecc} & 7B & 4.9 & 4.1 & 8.5 & 11.38 & 10.2 & 10.5 & 9.9 & 2.69 \\
\addlinespace[2pt]
Em-Garde~\cite{emgarde} & 2B+7B & 8.9 & 10.9 & 0.0 & 9.52 & 28.5 & 19.8 & 50.7 & 6.59 \\
\addlinespace[2pt]
Intern-Decision~\cite{intern} & 2B & 18.5 & 7.1 & 69.5 & 6.35 & 43.8 & 33.3 & 63.8 & 4.09 \\
\addlinespace[2pt]
OmniJev~\cite{omnijev} & 2B & 17.4 & 3.6 & \textbf{79.3} & 6.58 & 40.5 & 31.4 & 57.1 & 4.00 \\
\addlinespace[2pt]
Gemini 3.8 Live~\cite{geminilive} & \multicolumn{1}{c}{\na} & 20.5 & 10.9 & 63.4 & 4.19 & 16.9 & 29.5 & 11.9 & 0.91 \\
\addlinespace[2pt]
Qwen3.5-Omni-Flash-Realtime~\cite{qwenomnirealtime} & \multicolumn{1}{c}{\na} & 16.7 & 19.4 & 4.9 & 7.40 & 36.5 & 33.9 & 39.4 & 2.46 \\
\addlinespace[2pt]
Qwen3.8-Omni-Flash-Realtime~\cite{qwenomnirealtime} & \multicolumn{1}{c}{\na} & 28.8 & 34.7 & 2.4 & 4.68 & 40.4 & 39.5 & 41.4 & 2.04 \\
\midrule
\multicolumn{10}{@{}l}{\textbf{Fine-tuned}} \\
JoyAI-VL-Interaction~\cite{joyai} & 8B & 26.1 & 23.5 & 37.8 & 4.35 & 50.3 & 57.9 & 44.5 & 1.04 \\
\addlinespace[2pt]
MOSS-VL-Realtime~\cite{moss} & 11B & 26.3 & 20.5 & 52.4 & 4.65 & 52.3 & \textbf{69.2} & 42.1 & 0.60 \\
\addlinespace[2pt]
Intern-Decision~\cite{intern} & 2B & 18.3 & 7.9 & 64.6 & 6.41 & 56.0 & 47.3 & \textbf{68.6} & 2.45 \\
\addlinespace[2pt]
OmniJev~\cite{omnijev} & 2B & 15.8 & 3.0 & 73.2 & 7.13 & 44.1 & 37.6 & 53.3 & 2.84 \\
\addlinespace[2pt]
\noalign{\begingroup\color{black!8}\hrule height\normalbaselineskip\endgroup\vskip-\normalbaselineskip}
TEMPI & 2B+4B & \textbf{44.4} & \textbf{40.4} & 62.2 & \textbf{1.97} & \textbf{61.2} & 60.6 & 61.8 & 1.29 \\
\bottomrule
\end{tabular*}

\end{table*}

\subsection{Experimental Setup}
All methods are evaluated on the held-out test split of CoTriggerBench, comprising 448 task queries across 96 videos (Sec.~\ref{sec:benchmark}). We retain each baseline's official native inference pipeline and adapt its outputs to the notification protocol described below.

\paragraph{Baseline settings.}
The main comparison evaluates baselines through their full-instruction Direct interface and our complete TEMPI system. Controlled Direct/TEMPI comparisons are reported in the ablation studies. JoyAI-VL-Interaction~\cite{joyai} and MOSS-VL-Realtime~\cite{moss} retain native streaming memory at 1 fps, with non-silence responses counted as notifications. Intern-Decision~\cite{intern}, OmniJev~\cite{omnijev}, and our detector use four-frame windows at 5 fps. StreamingVLM~\cite{streamingvlm} and LiveCC~\cite{livecc} use native incremental inference at 2 fps with an \texttt{ALERT} prompt. Em-Garde~\cite{emgarde} uses native proposal--matching inference at 2 fps, with surge detections counted as notifications.

Gemini 3.8 Live~\cite{geminilive} and Qwen3.5/3.8-Omni-Flash-Realtime~\cite{qwenomnirealtime} use their native APIs with 1 fps video and \texttt{ALERT} readouts. Em-Garde is Direct-only; all other main-table baselines support paired Direct and TEMPI evaluation with backend weights and native sampling held fixed. 

\paragraph{Training settings.}
Our detector and all SFT baselines are fine-tuned on the CoTriggerBench training split to reduce mismatch with the evaluation domain. Baseline SFT follows the supervised fine-tuning formulations and output interfaces described in the official releases. Our detector uses the objective in Sec.~\ref{sec:detector}. Checkpoints, thresholds are selected on validation data and fixed for testing.

\paragraph{Metrics.}
All tasks are scored, including failed plans and missing outputs. As defined in Sec.~\ref{sec:benchmark}, TS measures whether the first notification falls within the accepted window on positive tasks, and AS measures whole-stream silence on adversarial tasks. Strict success, our primary metric, combines correct outcomes across both groups. MAE measures first-notification timing error over emitted positive tasks; silent tasks are excluded from MAE but remain scored in TS and Strict. The single-event columns report precision, recall, F1, and false event runs per minute (FP/min). 

\subsection{Baseline Comparison}
\paragraph{Overall performance.}
As shown in Table~\ref{tab:paired}, TEMPI achieves the highest strict success in the main comparison, 44.4\%, versus 28.8\% for Qwen3.8-Omni, the strongest baseline. Em-Garde attains a strict success of 8.9\% with no adversarial silence. For atomic event detection, our detector achieves the highest F1 (61.2\%), exceeding Intern-Decision (56.0\%) while producing fewer false events (1.29 versus 2.45/min). Across the four open-source baselines, fine-tuning improves precision and reduces false events.

\paragraph{Recognition--execution gap.}
Fine-tuning improves atomic-event F1 for all four open-source baselines, yet these gains do not consistently translate into Direct task success (Table~\ref{tab:paired}). Intern-Decision's F1 rises from 43.8\% to 56.0\% while strict success remains nearly unchanged (18.5\% to 18.3\%); OmniJev's F1 rises from 40.5\% to 44.1\% while strict success declines from 17.4\% to 15.8\%. With Direct thresholds selected on all 451 validation tasks, these classifiers achieve AS silence of 64.6\%--79.3\% but TS of only 3.0\%--7.9\%: withholding notifications reduces false alarms while missing most positive triggers. Stronger local recognition alone does not ensure correct execution of a compositional instruction. TEMPI maintains task progress and enforces the corresponding state transitions.

\subsection{Ablation Studies}
\paragraph{Effect of the TEMPI execution.}
We compare Direct and TEMPI execution while holding backend weights fixed. As shown in Table~\ref{tab:execution-ablation}, our workflow improves strict success for four of the five zero-shot backends shown and all four fine-tuned baselines. Explicit task decomposition and stateful execution improve overall success while exposing a trigger--silence tradeoff.
TEMPI-D denotes our 2B detector (D for detector). Direct execution supplies the full compositional instruction to TEMPI-D, whereas TEMPI execution uses its atomic-event predictions within our planning and scenario-state framework.
Table~\ref{tab:execution-ablation} shows that compiled execution more than doubles strict success from 18.3\% to 44.4\%; TS increases from 10.9\% to 40.4\%; AS rises from 51.2\% to 62.2\%.
The large gains highlight the value of decomposing compositional instructions and enforcing task-state constraints beyond event detection alone.

\begin{table}[t]
\centering
\caption{\textbf{Effect of TEMPI across perception backends.} D/T: Direct/TEMPI execution. D supplies the full instruction directly to the backend; T uses it as an atomic-event detector in our framework. TEMPI-D denotes our 2B detector. TS, AS, and Strict are percentages; MAE is in seconds. Better values within each pair are bold.}
\label{tab:execution-ablation}
\footnotesize
\setlength{\tabcolsep}{1pt}
\begin{tabular*}{\linewidth}{@{\extracolsep{\fill}}lrrrrrrrr@{}}
\toprule
\multirow{2}{*}{Backend} & \multicolumn{2}{c}{TS (\%) $\uparrow$} & \multicolumn{2}{c}{AS (\%) $\uparrow$} & \multicolumn{2}{c}{Strict (\%) $\uparrow$} & \multicolumn{2}{c}{MAE (s) $\downarrow$}  \\
\cmidrule(lr){2-3}\cmidrule(lr){4-5}\cmidrule(lr){6-7}\cmidrule(lr){8-9}
 & \multicolumn{1}{c}{D} & \multicolumn{1}{c}{T} & \multicolumn{1}{c}{D} & \multicolumn{1}{c}{T} & \multicolumn{1}{c}{D} & \multicolumn{1}{c}{T} & \multicolumn{1}{c}{D} & \multicolumn{1}{c}{T}  \\
\midrule
\multicolumn{9}{@{}l}{\textbf{Zero-shot}} \\
StreamingVLM & \textbf{3.8} & 0.8 & 68.3 & \textbf{96.3} & 15.6 & \textbf{18.3} & 6.39 & \textbf{4.51} \\
\addlinespace[1pt]
LiveCC & 4.1 & \textbf{9.3} & 8.5 & \textbf{47.6} & 4.9 & \textbf{16.3} & 11.38 & \textbf{6.99} \\
\addlinespace[1pt]
Gemini Live & \textbf{10.9} & 4.1 & 63.4 & \textbf{87.8} & \textbf{20.5} & 19.4 & \textbf{4.19} & 5.21 \\
\addlinespace[1pt]
Qwen3.5-Omni & 19.4 & \textbf{26.8} & 4.9 & \textbf{28.0} & 16.7 & \textbf{27.0} & 7.40 & \textbf{4.67} \\
\addlinespace[1pt]
Qwen3.8-Omni & \textbf{34.7} & 27.9 & 2.4 & \textbf{43.9} & 28.8 & \textbf{30.8} & 4.68 & \textbf{4.55} \\
\midrule
\multicolumn{9}{@{}l}{\textbf{Fine-tuned}} \\
JoyAI & 23.5 & \textbf{30.3} & 37.8 & \textbf{78.0} & 26.1 & \textbf{39.1} & 4.35 & \textbf{2.88} \\
\addlinespace[1pt]
MOSS & 20.5 & \textbf{26.8} & 52.4 & \textbf{78.0} & 26.3 & \textbf{36.2} & 4.65 & \textbf{3.17} \\
\addlinespace[1pt]
Intern-D & 7.9 & \textbf{44.5} & \textbf{64.6} & 36.6 & 18.3 & \textbf{43.1} & 6.41 & \textbf{2.79} \\
\addlinespace[1pt]
OmniJev & 3.0 & \textbf{23.0} & \textbf{73.2} & 40.2 & 15.8 & \textbf{26.1} & 7.13 & \textbf{4.03} \\
\addlinespace[1pt]
\noalign{\begingroup\color{black!8}\hrule height\normalbaselineskip\endgroup\vskip-\normalbaselineskip}
TEMPI-D & 10.9 & \textbf{40.4} & 51.2 & \textbf{62.2} & 18.3 & \textbf{44.4} & 5.25 & \textbf{1.97} \\
\bottomrule
\end{tabular*}

\end{table}

\begin{table}[t]
\centering
\caption{\textbf{Effect of input frame rate.} Better scores within each 1/5 fps pair are bold. Increasing the frame rate does not guarantee better performance; native settings yield higher TEMPI strict success and atomic-event F1 for both backends.}
\label{tab:fps-tasks}
\label{tab:fps-events}
\label{tab:input-density}
\footnotesize
\setlength{\tabcolsep}{0.8pt}
\begin{tabular*}{\linewidth}{@{\extracolsep{\fill}}lcrrrrrrrrr@{}}
\toprule
\multirow{2}{*}{Backend} & \multicolumn{1}{c}{\multirow{2}{*}{FPS}} & \multicolumn{3}{c}{Direct (\%)} & \multicolumn{3}{c}{TEMPI (\%)} & \multicolumn{3}{c}{Atomic events (\%)}  \\
\cmidrule(lr){3-5}\cmidrule(lr){6-8}\cmidrule(lr){9-11}
 &  & \multicolumn{1}{c}{TS$\uparrow$} & \multicolumn{1}{c}{AS$\uparrow$} & \multicolumn{1}{c}{Strict$\uparrow$} & \multicolumn{1}{c}{TS$\uparrow$} & \multicolumn{1}{c}{AS$\uparrow$} & \multicolumn{1}{c}{Strict$\uparrow$} & \multicolumn{1}{c}{P$\uparrow$} & \multicolumn{1}{c}{R$\uparrow$} & \multicolumn{1}{c}{F1$\uparrow$} \\
\midrule
JoyAI & 1 & \textbf{23.5} & 37.8 & 26.1 & \textbf{30.3} & \textbf{78.0} & \textbf{39.1} & \textbf{57.9} & \textbf{44.5} & \textbf{50.3} \\
JoyAI & 5 & 21.6 & \textbf{67.1} & \textbf{29.9} & 24.0 & 76.8 & 33.7 & 53.0 & 37.4 & 43.9 \\
\addlinespace[2pt]
MOSS & 1 & \textbf{20.5} & 52.4 & \textbf{26.3} & \textbf{26.8} & 78.0 & \textbf{36.2} & \textbf{69.2} & \textbf{42.1} & \textbf{52.3} \\
MOSS & 5 & 7.4 & \textbf{78.0} & 20.3 & 9.6 & \textbf{91.5} & 24.6 & 66.9 & 18.7 & 29.2 \\
\bottomrule
\end{tabular*}

\end{table}

\begin{table}[t]
\centering
\caption{\textbf{Scenario Planning Agent size.} Planning quality is reported as percentages. Qty.: correct subquery count rate. Sem.: position-wise semantic equivalence judged by Qwen3.8-Flash-Next-FP8~\cite{qwen38flashnext}. Match: correct count and equivalent queries. Type: task-type accuracy. Latency: complete-plan request time on one H800.}
\label{tab:planner-size}
\footnotesize
\setlength{\tabcolsep}{1pt}
\begin{tabular*}{\linewidth}{@{\extracolsep{\fill}}lrrrrrr@{}}
\toprule
\multirow{2}{*}{Planner} & \multicolumn{4}{c}{Planning quality (\%) $\uparrow$} & \multicolumn{2}{c}{Latency (ms) $\downarrow$}  \\
\cmidrule(lr){2-5}\cmidrule(lr){6-7}
 & \multicolumn{1}{c}{Qty.} & \multicolumn{1}{c}{Sem.} & \multicolumn{1}{c}{Match} & \multicolumn{1}{c}{Type} & \multicolumn{1}{c}{P50} & \multicolumn{1}{c}{P95} \\
\midrule
Qwen3.5-2B & 97.3 & 91.2 & 85.7 & 88.6 & 86.7 & 107.5 \\
\noalign{\begingroup\color{black!8}\hrule height\normalbaselineskip\endgroup\vskip-\normalbaselineskip}
Qwen3.5-4B & 99.3 & 92.7 & 87.7 & 90.9 & 129.0 & 160.6 \\
Qwen3.5-9B & 99.3 & 93.5 & 89.1 & 94.4 & 179.2 & 226.6 \\
Qwen3.6-35B-A3B & 98.7 & 93.4 & 88.6 & 98.2 & 159.6 & 244.2 \\
Qwen3.8-27B & 99.1 & 94.7 & 90.4 & 98.9 & 709.8 & 878.2 \\
\bottomrule
\end{tabular*}

\end{table}

\begin{table}[t]
\centering
\caption{\textbf{Detector GPU execution latency (ms).} Measurements are obtained on one H800 with a batch size of 1. TEMPI-D denotes our 2B detector.}
\label{tab:component-latency}
\small
\setlength{\tabcolsep}{2pt}
\begin{tabular}{@{}lcrrr@{}}
\toprule
Model & \multicolumn{1}{c}{Params} & \multicolumn{1}{c}{P50 $\downarrow$} & \multicolumn{1}{c}{P90 $\downarrow$} & \multicolumn{1}{c}{P99 $\downarrow$} \\
\midrule
\noalign{\begingroup\color{black!8}\hrule height\normalbaselineskip\endgroup\vskip-\normalbaselineskip}
TEMPI-D & 2B & 61.0 & 62.7 & 67.9 \\
Intern-Decision-2B~\cite{intern} & 2B & 75.2 & 82.0 & 85.7 \\
OmniJev-2B~\cite{omnijev} & 2B & 165.0 & 170.6 & 174.4 \\
JoyAI-VL-Interaction~\cite{joyai} & 8B & 240.3 & 469.6 & 573.1 \\
MOSS-VL-Realtime~\cite{moss} & 11B & 91.6 & 95.5 & 219.9 \\
StreamingVLM~\cite{streamingvlm} & 7B & 93.6 & 98.5 & 105.8 \\
LiveCC-7B-Instruct~\cite{livecc} & 7B & 67.8 & 84.4 & 116.9 \\
Em-Garde~\cite{emgarde} & 2B & 53.3 & 55.0 & 56.6 \\
\bottomrule
\end{tabular}

\end{table}


\paragraph{Frame-rate alignment.}
\label{sec:sampling-density}
We examine whether aligning baseline input frame rates with our detector improves performance. Fine-tuned JoyAI and MOSS receive 5 fps input instead of their native 1 fps, with consistent 1 s decision intervals, memory policies, and decoding settings fixed. 
As shown in Table~\ref{tab:input-density}, denser input reduces both TEMPI strict success and atomic-event F1 for both backends compared with their native settings.

\paragraph{Planner Capability.}

As shown in Table~\ref{tab:planner-size}, as the parameter quantity of the scenario planning agent increases, the quality of the scenario task memory construction improves accordingly, especially in task-type prediction and the strict match of sub-task decomposition and semantic consistency.
However, the most fundamental factor for the successful execution of a task is the consistency of the number of sub-tasks.
Beyond a parameter quantity of 4B, there is no significant improvement in this value.
The subtle differences in semantics may not lead to detection errors, and the classification error of task type may not affect the execution sequence of the execution graph.
To balance inference speed and effectiveness, we select the 4B agent as the final scenario planning agent.

\paragraph{Computational Efficiency.}

TEMPI requires only one planning call per instruction, leaving repeated inference of the event detector.
The planner incurs a latency of 129 ms on an H800 (Table~\ref{tab:planner-size}).
For subsequent detection, our detector takes 61.0 ms per query/window, compared with 240.3 ms for JoyAI, using one-quarter of the parameters with a $3.94\times$ speedup in GPU execution (Table~\ref{tab:component-latency}).
Under the same TEMPI framework, it also achieves higher strict success than JoyAI (44.4\% vs.\ 39.1\%; Table~\ref{tab:execution-ablation}), demonstrating efficient continuous detection with strong task performance.

\subsection{Edge Deployment}

\begin{figure}[t]
\centering
\begin{subfigure}{\linewidth}
\includegraphics[width=\linewidth]{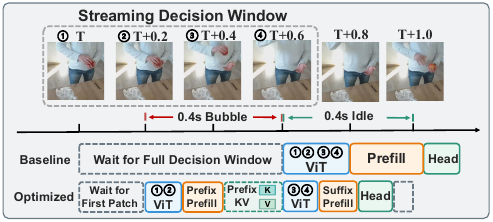}
\caption{Process flow diagram for early prefix execution}\label{fig:rtx5060-prefix}
\end{subfigure}\par\vspace{2pt}
\begin{subfigure}{\linewidth}
\includegraphics[width=\linewidth]{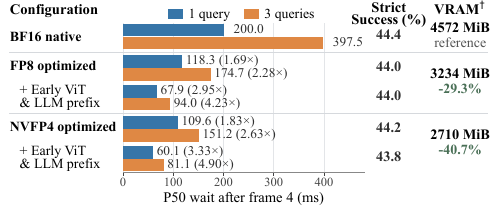}
\caption{Results for different configurations on RTX 5060}\label{fig:rtx5060-comparison}
\end{subfigure}
\caption{\textbf{Local efficiency and early execution on an RTX 5060 (8 GB).} Figure~\subref{fig:rtx5060-prefix} shows how early ViT and prefix prefill overlap frame arrival; cached KV feeds suffix prefill after frame 4. Figure~\subref{fig:rtx5060-comparison} reports decision latency P50 for one/three queries and independently evaluated strict success. $\dagger$: VRAM measured per precision with three queries.}
\label{fig:rtx5060-efficiency}
\end{figure}

\paragraph{Streaming Prefix Overlap.}
We exploit idle intervals in streaming frame collection to overlap computation with frame arrival.
In the baseline pipeline, inference starts only after all four frames in a decision window arrive.
The wait for frames 3--4 therefore creates a computation bubble, where frames 1--2 are already available, but their processing is deferred.
In Figure~\ref{fig:rtx5060-prefix}, visual encoding of frames 1--2 is independent of frames 3--4, allowing us to fill this bubble with early ViT encoding and LLM prefix prefill.
Once frame 4 arrives, inference continues with the remaining frames and suffix prefill, reusing the prefix KV caches.
This remaining computation fits within the otherwise idle interval in which the next window collects frames 1--2, completing before its prefix computation begins.
The schedule thus reduces decision latency after the final frame arrives while avoiding contention between consecutive windows.

\paragraph{Result Analysis.}

We deploy the detector on a single RTX 5060 with 8 GB of VRAM.
In addition to streaming prefix overlap, we apply quantization to reduce memory consumption.
The FP8 implementation uses native DeepGEMM~\cite{deepgemm} kernels, while the NVFP4 implementation uses native cuBLASLt~\cite{cublaslt} FP4 GEMMs with fused scaling and output operations.
We further reduce execution overhead through compilation, kernel fusion, CUDA graphs, and shared activation quantization.
As shown in Figure~\ref{fig:rtx5060-comparison}, the FP8 configuration with early execution achieves $2.95\times$ and $4.23\times$ speedups over native BF16 in median decision latency for one and three queries per batch, respectively.
The NVFP4 configuration with early execution increases these speedups to $3.33\times$ and $4.90\times$.
Separate per-precision measurements show VRAM reductions of 29.3\% and 40.7\% for FP8 and NVFP4, respectively.

\section{Conclusion}

We introduce CoTriggerBench to evaluate compositional event triggering in streaming video and TEMPI to bridge the gap between event recognition and instruction execution. By maintaining persistent task progress through Scenario Memory, TEMPI achieves state-of-the-art instruction-level success (44.4\%), while its compact 2B detector achieves the highest atomic-event F1 (61.2 \%) among evaluated models. Furthermore, we combine low-bit quantization with early ViT and LLM prefix execution to optimize streaming inference, achieving up to a 4.90$\times$ speedup in post-window decision latency on an 8 GB RTX 5060.

\label{main-text-end}
\section*{LLM Usage Statement}
\label{llm-usage-start}
DeepSeek-V4.1-Flash was used to generate natural-language benchmark instructions. GLM-5.3-Flash and Gemini 3.8 Flash assisted with visual-grounding checks during benchmark construction. Qwen3.8-Flash-Next-FP8 was used to judge the semantic equivalence of planner-generated atomic queries. Qwen3.5-4B and Qwen3-VL-2B-Instruct provide the planning and perception components described in Sec.~\ref{sec:method}; other evaluated LLM-based systems are identified in Sec.~\ref{sec:experiments}. OpenAI Codex assisted with manuscript editing, language polishing, LaTeX formatting, and submission checks. The authors retain responsibility for the scientific content, experimental results, citations, and conclusions.

{\small
\bibliographystyle{ieeenat_fullname}
\bibliography{main}
}
\end{document}